\documentclass[runningheads,a4paper]{llncs}
\usepackage[T1]{fontenc}
\usepackage{graphicx}
\usepackage{subcaption} % defines the subfigure environment
\usepackage{float}      % helps with [H] placement
\usepackage{microtype}
\usepackage{amsmath,amssymb}

\usepackage[table]{xcolor} % for \rowcolor
\usepackage{booktabs}      % for \cmidrule and other table features
\usepackage{adjustbox}     % for adjustbox environment
\usepackage{threeparttable} % for tablenotes environment

\usepackage{enumitem}
\setlist{nosep, leftmargin=14pt}

\usepackage{color}
\usepackage[final]{hyperref} % changed from [draft] to [final]

\hypersetup{pdfborder={0 0 0}}

\begin{document}
\title{MMRINet: Efficient Mamba-Based Segmentation with Dual-Path Refinement for Low-Resource MRI Analysis}
\titlerunning{MMRINet: Efficient Mamba-Based Segmentation for Low-Resource MRI}

%
%\titlerunning{Abbreviated paper title}
% If the paper title is too long for the running head, you can set
% an abbreviated paper title here
%
\author{Abdelrahman Elsayed\thanks{Equal contribution; names are presented alphabetically}\inst{1} \and
Ahmed Jaheen$^{\star}$\inst{1,2} \and
Mohammad Yaqub\inst{1}}
 
\authorrunning{A. Elsayed, A. Jaheen and M. Yaqub}
 
\institute{Mohamed bin Zayed University of Artificial Intelligence (MBZUAI), Abu Dhabi, United Arab Emirates\\
\and
New York University (NYU), New York, USA\\
\email{\{abdelrahman.elsayed, ahmed.jaheen, mohammad.yaqub\}@mbzuai.ac.ae}\\
\email{aj3379@nyu.edu}}

\maketitle              % typeset the header of the contribution

\begin{abstract}
Automated brain tumor segmentation in multi-parametric MRI remains a critical yet underserved challenge in resource-constrained clinical settings, where deep 3D networks requiring high-end GPUs are not viable. This is particularly acute across sub-Saharan Africa (SSA), where low-field scanners, heterogeneous patient demographics, and severe data scarcity compound the difficulty of applying standard deep learning pipelines. We present \textbf{MMRINet}, a lightweight segmentation architecture purpose-built for these constraints. At its core, MMRINet replaces quadratic-complexity self-attention with linear-complexity Mamba state-space models~\cite{mamba}, enabling efficient long-range volumetric context modeling without the computational overhead of Transformer-based approaches. We combine two lightweight refinement components: \textbf{Dual-Path Feature Refinement (DPFR)}, which extracts complementary detail and contextual representations to improve feature diversity under limited data, and \textbf{Progressive Feature Aggregation (PFA)}, which hierarchically fuses multi-scale decoder outputs for sharper segmentation boundaries. Evaluated on the BraTS-Lighthouse SSA 2025 challenge dataset~\cite{bratsafrica2025}, comprising 3D MRI scans from Nigerian clinical sites, MMRINet achieves an average Dice score of \textbf{0.752} and an average HD95 of \textbf{12.23\,mm} with only \textbf{$\sim$2.5M} parameters, outperforming all evaluated baselines, including UNETR, Swin-UNETR, SegMamba, and SegResNet3D. These results indicate that strong validation-set segmentation performance can be achieved with substantially reduced computation, offering a practical step toward AI-assisted neuro-oncology in low-resource clinical environments. %Our GitHub repository can be accessed here: github.com/BioMedIA-MBZUAI/MMRINet.

\end{abstract}
\section{Introduction}
\label{sec:intro}
Brain tumor segmentation from multi-parametric MRI is a foundational task in neuro-oncology, supporting diagnosis, treatment planning, response assessment, and longitudinal monitoring of gliomas~\cite{wen2008malignant,louis2021who,menze2015brats}. Manual segmentation by trained radiologists is accurate but time-consuming, prone to inter-observer variability, and infeasible at scale in regions facing a critical shortage of specialist clinicians. Hence, deep learning has emerged as a promising path toward automated, reproducible tumor delineation~\cite{surveyseg}. Standard deep learning solutions for this task, most prominently the family of 3D U-Net architectures~\cite{3dunet,vnet,nnunet}, have achieved strong benchmark performance but carry heavy computational demands. These models typically require high-memory GPUs, large annotated training cohorts, and significant engineering overhead to deploy reliably. While suitable for well-resourced academic medical centres, they are difficult to translate to the clinical settings of sub-Saharan Africa (SSA), where MRI infrastructure is frequently constrained, image quality is variable, and annotated data is scarce.
 
Transformer-based architectures such as UNETR~\cite{unetr} and Swin-UNETR~\cite{swinunetr} introduced self-attention mechanisms to capture long-range spatial dependencies, achieving strong performance on Western benchmark datasets including multiple BraTS editions~\cite{bratsafrica2025}. However, self-attention scales quadratically with input size, rendering these models computationally expensive at the volumetric resolutions needed for 3D MRI analysis. Moreover, their performance degrades substantially when applied to SSA cohorts~\cite{bratsafrica}, where scanner variability, acquisition noise, and limited training examples undermine the distributional assumptions built into models pretrained on high-field Western data.
 
A growing body of work has sought to close this gap through architectural efficiency improvements. Convolutional variants such as Attention U-Net~\cite{oktay2018attention}, UNet++~\cite{zhou2018unet++}, and Efficient-UNet~\cite{uzen2023depth} introduced hierarchical skip connections, compound scaling, and parameter pruning to reduce computational cost. Domain adaptation and self-supervised learning approaches~\cite{perone2019unsupervised,li2020domain,zhou2022models} have addressed the data scarcity dimension. Parameter-efficient fine-tuning strategies~\cite{parida2024adult,hashmi2024optimizing,jaheen2025emednext} have also demonstrated partial success in adapting large pretrained models to SSA-specific datasets. However, the fundamental tension remains: existing architectures are often too heavy, too data-hungry, or too reliant on high-quality pretraining to be practically deployable in low-resource environments.
 
Recent state-space models (SSMs) offer a principled resolution to this tension. Mamba~\cite{mamba} achieves linear complexity in sequence length through selective state space dynamics, enabling efficient long-range dependency modeling without the quadratic cost of attention. Medical imaging adaptations including U-Mamba~\cite{umamba} and SegMamba~\cite{segmamba} have demonstrated that SSM blocks can be integrated within U-Net-like encoder-decoder architectures to achieve competitive 3D segmentation performance. Group-wise variants~\cite{groupmamba} further improve parameter efficiency by partitioning channels across directional scanners, but their extension to volumetric MRI requires explicit handling of 3D spatial axes and anisotropic scan order. Existing Mamba-based medical imaging models have not been designed with the specific constraints of SSA clinical deployment in mind: they achieve efficiency primarily through architectural substitution (replacing attention with SSMs) without addressing the complementary challenges of feature diversity under extreme data scarcity or robust multi-scale fusion for challenging, noisy volumetric inputs. The BraTS-Lighthouse SSA 2025 challenge~\cite{bratsafrica2025} has established a dedicated benchmark for precisely this setting, and recent competitive entries~\cite{parida2024adult,hashmi2024optimizing,jaheen2025emednext} remain parameter-heavy and largely reliant on external pretraining.
 
We address this gap with \textbf{MMRINet}, a lightweight architecture explicitly designed for low-resource 3D brain tumor segmentation. MMRINet makes three targeted system-level contributions:
 
\begin{enumerate}
    \item \textbf{Modulated GroupMamba Bottleneck}: We introduce a 3D volumetric extension of GroupMamba for MRI segmentation, using group-wise, multi-directional state-space scanning across depth, height, and width axes to preserve 3D spatial coherence while modeling global context with $1/G$ of the parameters of full-channel scanning, where $G$ denotes the number of channel groups.
 
    \item \textbf{Dual-Path Feature Refinement (DPFR)}: A lightweight decoder module that simultaneously extracts fine-grained detail (via depthwise-separable convolution) and broad contextual patterns (via dilated convolution), fusing them through adaptive gating to improve feature diversity without requiring additional data or augmentation.
 
    \item \textbf{Progressive Feature Aggregation (PFA)}: A hierarchical multi-scale fusion strategy that aggregates decoder features across resolution levels, producing segmentation outputs with consistently strong boundary precision.
\end{enumerate}

\noindent
With deep supervision and only \textbf{$\sim$2.5M} parameters, MMRINet achieves the best average Dice \textbf{(0.752)} and the best average HD95 \textbf{(12.23\,mm)} among all evaluated baselines on the \textbf{BraTS-Lighthouse SSA 2025 validation set}, while requiring substantially less memory and compute than any Transformer-based competitor. We focus on this dataset because MMRINet is targeted at a specific, underserved clinical context, and the SSA dataset presents unique challenges of scanner noise, morphological variability, and data scarcity.

\section{Related Work}
\label{sec:related}
 
\subsection{Volumetric Brain Tumor Segmentation}
 
The 3D U-Net~\cite{3dunet} established the encoder-decoder paradigm for volumetric medical image segmentation, using skip connections to preserve spatial detail across resolution scales. Subsequent work introduced residual encoders (V-Net~\cite{vnet}), self-configuring training pipelines (nnU-Net~\cite{nnunet}), and hierarchical architectural improvements (UNet++~\cite{zhou2018unet++}, Attention U-Net~\cite{oktay2018attention}). These models perform well on high-quality benchmark datasets but can be sensitive to distributional shift, scanner variability, and limited training data~\cite{bratsafrica,perone2019unsupervised,zhou2022models}.
 
Transformer-based methods~\cite{unetr,swinunetr} introduced global context modeling via self-attention, achieving strong results at the cost of quadratic computational complexity. While effective for large-scale datasets, their compute demands and sensitivity to data volume make them poorly suited to SSA settings. SegResNet3D~\cite{myronenko20183d} offered a middle ground through autoencoder regularization, but still requires substantial GPU memory for volumetric inputs.
 
\subsection{Mamba and State-Space Models in Medical Imaging}
 
The Mamba architecture~\cite{mamba} introduces selective state-space dynamics that achieve linear complexity in sequence length, making it particularly attractive for long-sequence and high-resolution 3D inputs. U-Mamba~\cite{umamba} first demonstrated that Mamba blocks can replace attention in biomedical segmentation without loss of performance. SegMamba~\cite{segmamba} extended this to 3D MRI segmentation with competitive BraTS results. GroupMamba~\cite{groupmamba} refined parameter efficiency through channel group partitioning and multi-directional scanning in visual state-space modeling.
 
A key concern for Mamba-based 3D models is the treatment of spatial structure: sequential scanning of 3D volumes introduces anisotropy as spatial relationships along the scanning axis are treated differently from transverse ones. MMRINet addresses this by extending GroupMamba to volumetric MRI through multi-directional group scanning across all three spatial axes, explicitly mitigating this limitation at the architectural level.

\subsection{Decoder Refinement and Multi-Scale Fusion}

MMRINet's decoder design is related to multi-branch and multi-scale refinement modules used in prior segmentation networks. ASPP-style dilated convolution modules capture wider context through parallel receptive fields~\cite{deeplabv3}, while Attention U-Net and gated fusion mechanisms learn to emphasize task-relevant features~\cite{oktay2018attention}. UNet++ uses nested skip pathways to improve multi-scale feature reuse~\cite{zhou2018unet++}. DPFR and PFA are therefore best understood as a compact integration of these established principles for the low-data SSA setting: DPFR uses only two low-cost 3D branches to balance local detail and dilated context after skip fusion, and PFA performs output-side aggregation of decoder levels without adding a heavy feature pyramid.
 
\subsection{Brain Tumor Segmentation in Sub-Saharan Africa}
 
The BraTS-Africa benchmark~\cite{bratsafrica} and its successor BraTS-Lighthouse SSA 2025~\cite{bratsafrica2025} have established dedicated evaluation protocols for glioma segmentation in Nigerian clinical populations. These datasets differ substantially from Western benchmarks: they feature lower field strength (resulting in higher noise and lower contrast), greater demographic variability, and far fewer training cases. Models pretrained on Western data and fine-tuned on SSA data have shown performance gaps~\cite{parida2024adult}, and parameter-heavy architectures designed for high-resource settings~\cite{hashmi2024optimizing,jaheen2025emednext} require adaptation strategies that may not translate to direct clinical deployment. MMRINet is the first architecture designed from the ground up for SSA-specific constraints rather than adapted from general-purpose models.

\section{Methods}
\label{sec:methods}
 
\subsection{Dataset}
\label{subsec:dataset}
All experiments are conducted on the \textbf{BraTS-Lighthouse SSA 2025} dataset \cite{bratsafrica2025,bratsafrica}, which constitutes the only publicly available 3D multi-parametric MRI benchmark collected from sub-Saharan African clinical sites, primarily hospitals in Nigeria. Each subject includes the four standard MRI modalities: T1, T1-contrast-enhanced (T1ce), T2, and FLAIR, accompanied by expert-annotated tumor masks delineating three subregions: Enhancing Tumor (ET), Tumor Core (TC), and Whole Tumor (WT).
 
\begin{figure}
    \centering
    \begingroup
    \setlength{\tabcolsep}{2pt}
    \begin{tabular}{ccc}
    \includegraphics[width=0.31\linewidth,trim={0pt 0pt 576.15pt 0pt},clip]{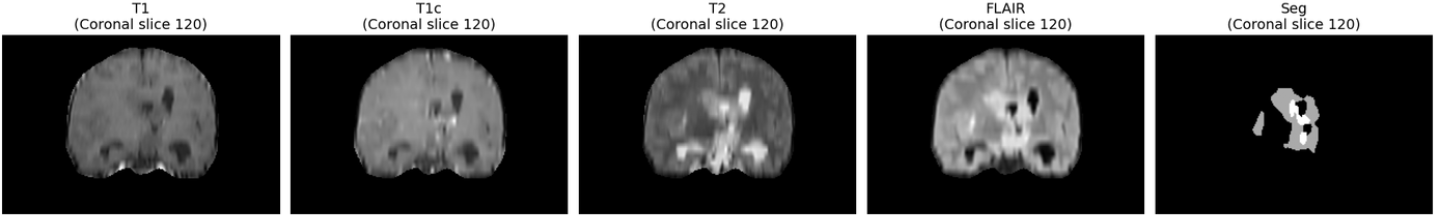} &
    \includegraphics[width=0.31\linewidth,trim={144.04pt 0pt 432.11pt 0pt},clip]{SSA.png} &
    \includegraphics[width=0.31\linewidth,trim={288.08pt 0pt 288.08pt 0pt},clip]{SSA.png} \\
    \multicolumn{3}{c}{%
    \includegraphics[width=0.31\linewidth,trim={432.11pt 0pt 144.04pt 0pt},clip]{SSA.png}\hspace{4pt}%
    \includegraphics[width=0.31\linewidth,trim={576.15pt 0pt 0pt 0pt},clip]{SSA.png}}
    \end{tabular}
    \endgroup
    \caption{Cross sections of the four modalities obtained from a sample data-point from the SSA dataset along with the corresponding segmentation masks}
    \label{fig:ssa-example}
\end{figure}

The dataset consists of 60 training cases and 35 validation cases. It is characterized by substantially higher noise levels, lower image contrast, and greater morphological variability than Western counterparts, reflecting the reality of low-field scanner acquisition at resource-limited clinical sites. These properties make it an ideal and clinically meaningful benchmark for evaluating segmentation systems intended for SSA deployment. An example case is shown in Figure~\ref{fig:ssa-example}.

The challenge scenarios presented by SSA MRI, field heterogeneity, limited training data, and acquisition artifacts, are distinct from those in standard benchmarks. Demonstrating strong performance in this specific setting is the primary goal of this work. We therefore report results on the fixed challenge validation split and discuss cross-dataset generalization as a limitation and future direction rather than inferring it from this benchmark alone.

\subsection{Data Preprocessing and Augmentation}
\label{subsec:preprocessing}

All MRI volumes undergo standardized preprocessing prior to training. Intensities are normalized per modality using statistics computed from non-zero voxels (i.e., within the brain mask), ensuring that background regions do not skew channel-wise normalization. No skull stripping or registration beyond what is provided in the challenge dataset is applied.
 
During training, random spatial crops of size $160 \times 160 \times 128$ voxels are extracted to accommodate GPU memory constraints while exposing the model to diverse volumetric contexts. This crop-based strategy also acts as a form of implicit data augmentation, as each training step presents a different sub-volume.
 
To improve robustness to the scanner variability characteristic of SSA MRI, we apply a targeted augmentation pipeline: (i) random spatial flipping along all three axes independently, each with probability 0.5; (ii) random intensity scaling by a factor uniformly drawn from $[0.9, 1.1]$; and (iii) random intensity shifting by an additive offset drawn from $[-0.1, 0.1]$. These transformations introduce realistic intensity variation while preserving anatomical structure and segmentation label validity, directly mimicking the kind of acquisition-level variability seen across scanner types in the SSA dataset.

\subsection{Network Architecture}
\label{sec:architecture}
\begin{figure}[!t]
    \centering
    \includegraphics[width=0.84\linewidth]{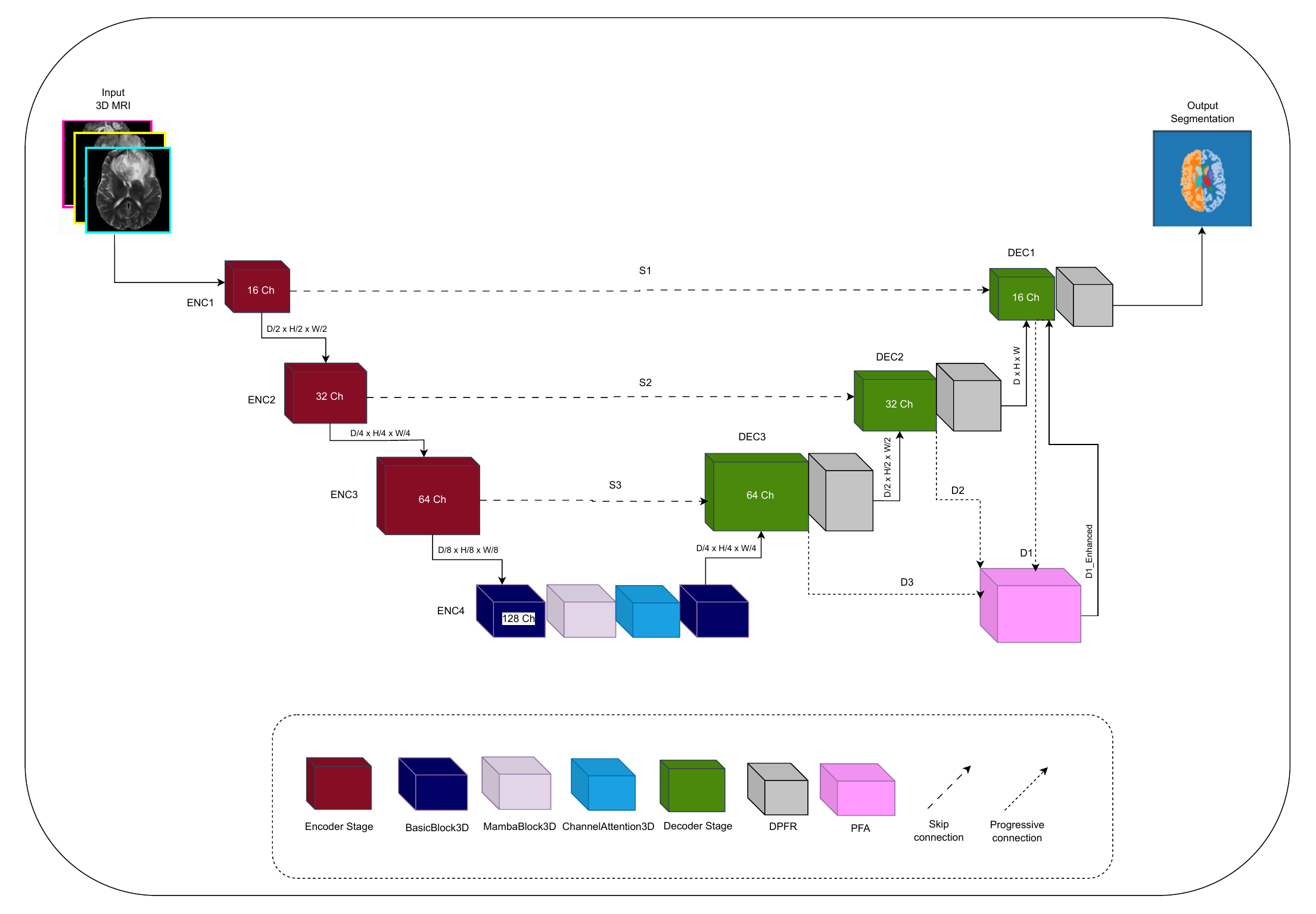}
    \caption{Overview of the MMRINet architecture. A residual convolutional encoder feeds into a Mamba-based bottleneck for efficient global context modeling. Three decoder stages with Dual-Path Feature Refinement (DPFR) progressively upsample and refine features, which are hierarchically aggregated via Progressive Feature Aggregation (PFA) to produce the final segmentation.}
    \label{fig:mmrinet-arch}
\end{figure}

MMRINet follows a U-Net encoder-decoder structure optimized for low-resource 3D medical image segmentation. Key design components include: (i) Mamba-based bottleneck for efficient global modeling, (ii) Dual-Path Feature Refinement (DPFR) in decoders, and (iii) Progressive Feature Aggregation (PFA). The architecture overview is shown in Figure~\ref{fig:mmrinet-arch}.

\subsubsection{Encoder.}
The encoder comprises four stages of \textit{BasicBlock3D} residual units with strided downsampling. Beginning from $C_\text{in}=4$ input channels (one per modality), the feature map dimensionality expands to $16$, $32$, $64$, and $128$ channels through successive stages, while spatial resolution reduces by a factor of two per stage along each spatial dimension. Each BasicBlock3D employs two $3\times3\times3$ convolutions with batch normalization, ReLU activation, and dropout ($p=0.05$). Skip connections from each encoder stage are preserved for decoder fusion.

\subsubsection{Mamba Bottleneck.}
At the coarsest encoder resolution ($D/8 \times H/8 \times W/8$, 128 channels), the bottleneck captures global spatial context via three sequential components:
 
\noindent\textbf{(1) GroupMamba3D}~\cite{groupmamba}: We extend the group-wise visual state-space formulation to 3D MRI volumes. Channel features are partitioned into $G=4$ groups with state-space dimension $d_\text{ssm}=32$, and each group independently applies directional scanning along the $H$, $W$, $D$, and diagonal $HW$ axes, enabling parallel capture of volumetric dependencies. Within each group, the processing pipeline consists of depthwise convolution, SSM projection, and layer normalization, followed by cross-group modulation to integrate information across partitions. Multi-directional scanning is the key mechanism by which MMRINet avoids the anisotropy artifacts associated with single-axis sequential Mamba scanning~\cite{mamba}. By scanning across all principal spatial axes within each group, the model treats volumetric relationships symmetrically. This yields an effective global receptive field at $1/G$ the parameter cost of full-channel Mamba, making it viable for $128^3$ volume processing on a single GPU.
 
\noindent\textbf{(2) Channel Attention}: A squeeze-excitation module (reduction ratio $r=8$) recalibrates feature channels by explicitly modeling inter-channel dependencies, amplifying task-relevant activations and suppressing redundant ones.
 
\noindent\textbf{(3) BasicBlock3D}: A final residual refinement block restores spatial detail that may be softened by the SSM projection step.

\begin{figure}[!t]
    \centering
    \includegraphics[width=0.84\linewidth]{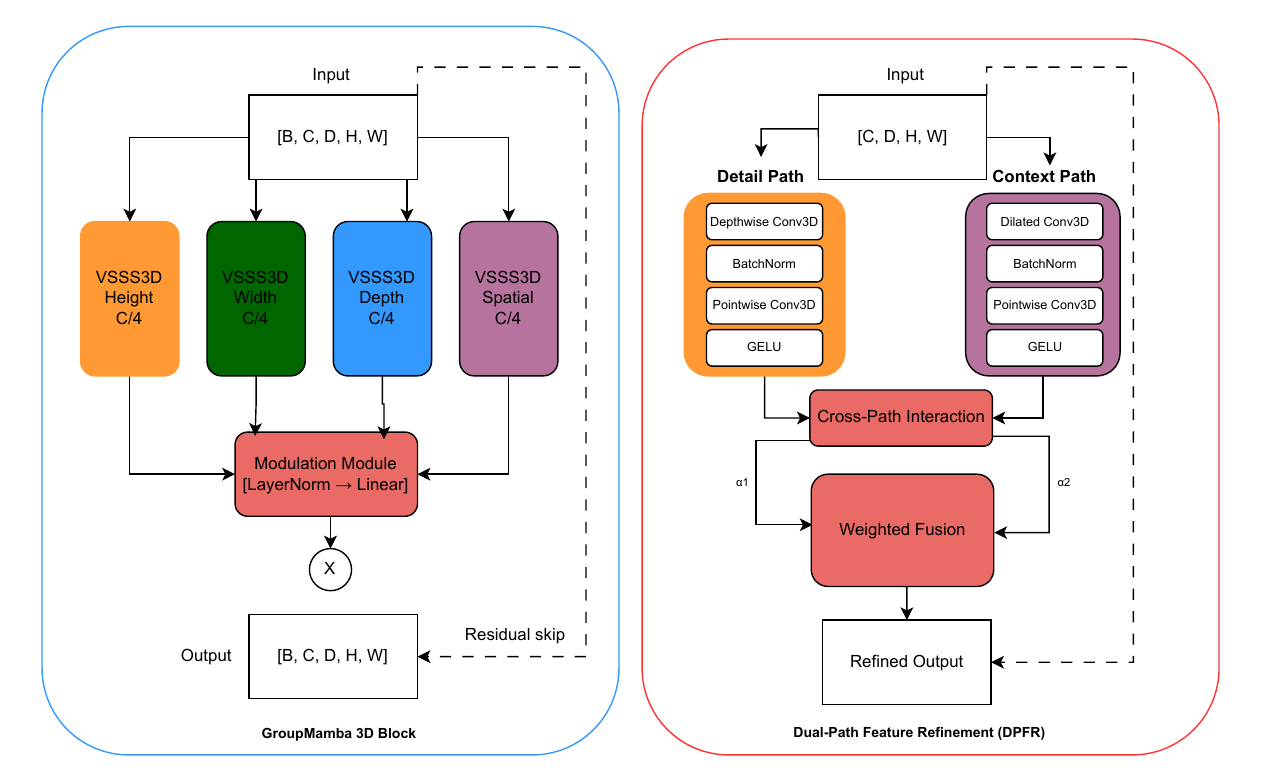}
    \caption{Key architectural components of MMRINet. \textbf{(Left)} The GroupMamba 3D Block models long-range spatial dependencies via group-wise state-space mixing with multi-directional scanning, maintaining 3D spatial coherence at linear complexity. \textbf{(Right)} The Dual-Path Feature Refinement (DPFR) module combines fine-grained detail (depthwise-separable conv) and contextual features (dilated conv) through adaptive gating, maximizing representational diversity within each decoder stage.}
    \label{fig:mamba-dpfr}
\end{figure}

\subsubsection{Decoder with Dual-Path Feature Refinement (DPFR).}
Three decoder stages progressively upsample feature maps via learned transposed convolution, concatenate corresponding encoder skip connections, and refine the merged representation via BasicBlock3D. We retain transposed convolutions to preserve learnable upsampling capacity while keeping the full model at only 2.5M parameters. Each decoder stage additionally incorporates the \textbf{DPFR module}, illustrated in Figure~\ref{fig:mamba-dpfr}:
 
\begin{itemize}
    \item \textbf{Detail Path}: A $3\times3\times3$ depthwise-separable convolution preserves high-frequency spatial detail, including fine tumor boundaries and small enhancing lesion regions that are particularly important for HD95 performance.
    \item \textbf{Context Path}: A dilated $3\times3\times3$ convolution (dilation $= 2$, effective receptive field $= 5\times5\times5$) captures broader contextual patterns without increasing parameter count.
    \item \textbf{Cross-Path Interaction}: The two paths are concatenated and processed via a $1\times1\times1$ convolution to enable information exchange between detail and context representations.
    \item \textbf{Adaptive Gating}: A lightweight gating network, implemented via global average pooling followed by a two-layer MLP and softmax normalization, learns to balance the contribution of each path based on the input feature statistics. This allows the model to emphasize fine detail for small ET regions and broader context for diffuse WT boundaries.
\end{itemize}
DPFR functions as an implicit ensemble within each decoder stage, maximizing representational diversity without requiring additional training data or auxiliary inputs. The dual-path design is directly motivated by the SSA dataset's challenges: heterogeneous tumor morphologies and high noise require a model that can simultaneously detect fine structural detail and suppress noisy background activations via contextual grounding.

\subsubsection{Progressive Feature Aggregation and Output Head.}
\textbf{PFA} hierarchically aggregates multi-scale decoder features:
\begin{equation}
\mathcal{F}_\text{agg} = \mathcal{U}\!\left(\mathcal{U}(\mathcal{D}_3) + \mathcal{D}_2\right) + \mathcal{D}_1,
\end{equation}
where $\mathcal{U}$ denotes trilinear upsampling with channel alignment when required, and $\mathcal{D}_i$ denotes the output of the $i$-th decoder stage. The aggregated multi-scale representation $\mathcal{F}_\text{agg}$ is combined with $\mathcal{D}_1$ via a residual connection, refined through a BasicBlock3D, and projected to $C_\text{out}=3$ output channels (ET, TC, WT) via a $1\times1\times1$ convolution.
 
\noindent\textbf{Deep Supervision}: During training, auxiliary segmentation heads are attached to intermediate decoder stages $\mathcal{D}_1$, $\mathcal{D}_2$, and $\mathcal{D}_3$. These heads are removed at inference time, contributing no additional cost. The inclusion of auxiliary supervision improves gradient flow to early decoder layers, which is particularly beneficial in compact architectures where the gradient signal from the final output head alone may be insufficient.

\subsection{Loss Function}
\label{subsec:loss}
We employ a combined Dice-Focal loss to address the severe class imbalance inherent in brain tumor segmentation, where tumor voxels typically constitute less than 5\% of the total brain volume. The segmentation loss is:
\begin{equation}
\mathcal{L}_\text{seg} = \mathcal{L}_\text{Dice} + \mathcal{L}_\text{Focal}.
\end{equation}
The Dice loss~\cite{dsc} encourages high overlap between predicted and ground-truth masks at a region level, while the Focal loss~\cite{focalloss} reweights per-voxel binary cross-entropy to emphasize hard, misclassified examples. We set the focal modulation factor $\gamma = 2.0$, which provides strong emphasis on difficult samples without destabilizing training. Predictions are obtained via sigmoid activation rather than softmax because the BraTS target regions are overlapping composite labels (ET, TC, WT). At inference, all channels use a fixed threshold of 0.5; no per-class thresholds are tuned on the validation set.
 
With deep supervision applied at three intermediate decoder stages, the total training loss is a weighted combination:
\begin{equation}
\mathcal{L} = 0.2\,\mathcal{L}(\text{out}_1) + 0.3\,\mathcal{L}(\text{out}_2) + 0.3\,\mathcal{L}(\text{out}_3) + 0.2\,\mathcal{L}(\text{final}).
\end{equation}
The moderate weighting on intermediate outputs encourages meaningful representations at each resolution level, while the balanced final output weight prevents the main prediction head from being overshadowed. In inference, only the final output is used.

\section{Experimental Setup}
\label{sec:experiments}

\subsection{Training Configuration}
In keeping with our low-resource design philosophy, MMRINet was trained on a single NVIDIA RTX A6000 GPU using mixed-precision (FP16) arithmetic to minimize memory and computation overhead. Training converged within 100 epochs. We used the AdamW optimizer with an initial learning rate of $3\times10^{-4}$ and weight decay of $1\times10^{-2}$. Rather than a manually tuned learning rate schedule, we employed a schedule-free learning rate strategy~\cite{schedulefree} ($r=1.0$, \texttt{weight\_lr\_power}$=0.5$). We use this primarily as a practical training choice and do not claim an isolated improvement over cosine or polynomial schedules.
 
The batch size was set to 2 volumes per step, reflecting the memory constraints imposed by large crop sizes ($160\times160\times128$). Despite this small batch size, training remained stable throughout (Section~\ref{subsec:loss}), a property we attribute to the combination of batch normalization and deep supervision that together regularize gradient flow at multiple levels of the network.

\subsection{Baseline Models}
We compare MMRINet against five established volumetric segmentation architectures: UNETR~\cite{unetr}, Swin-UNETR~\cite{swinunetr}, SegMamba~\cite{segmamba}, SegResNet3D~\cite{myronenko20183d}, and MONAI 3D U-Net~\cite{3dunet}. All baselines are implemented using the MONAI framework~\cite{monai} with settings adapted to the small SSA dataset to minimize overfitting (e.g., reduced initial feature channels, increased dropout, conservative learning rates). SegMamba uses the original implementation with equivalent tuning. All models are trained under identical hardware constraints, input crop size, mixed-precision setting, and augmentation pipeline. This comparison is intentionally architecture-focused and resource-controlled: it contrasts compact 3D segmentation networks under the same memory and training protocol rather than evaluating full auto-configuration pipelines such as nnU-Net~\cite{nnunet}, whose search and preprocessing policies target a different experimental scope.

\subsection{Evaluation Metrics}
\label{sec:metrics}
Performance is evaluated using the Dice Similarity Coefficient (DSC) and 95th-percentile Hausdorff Distance (HD95) for three tumor subregions: Enhancing Tumor (ET), Tumor Core (TC), and Whole Tumor (WT). Overall performance is summarized as the mean DSC and HD95 across all three subregions.

\section{Results}
\label{sec:results}

\subsection{Quantitative Evaluation}
Table~\ref{tab:results} presents segmentation performance on the BraTS-Lighthouse SSA 2025 validation set. MMRINet achieves an average Dice of \textbf{0.752} and an average HD95 of \textbf{12.23\,mm}, outperforming all evaluated baselines on both average metrics. The main exception at subregion level is ET HD95: the MONAI 3D U-Net achieves a lower ET HD95 (9.57\,mm vs. 16.21\,mm), reflecting stronger boundary precision for the enhancing tumor subregion specifically. However, this comes at the cost of substantially poorer TC and WT HD95 (22.13\,mm and 21.95\,mm, respectively), yielding a worse average HD95 of 17.88\,mm compared to MMRINet's 12.23\,mm.
 
MMRINet's strongest relative performance is in the Whole Tumor subregion (0.84 DSC), reflecting robust detection of the outer tumor boundary. Enhancing Tumor remains the most challenging subregion (0.69 DSC), consistent with the small, heterogeneous nature of ET in the SSA cohort. The Tumor Core performance (0.73 DSC, 8.12\,mm HD95) shows particularly strong boundary precision, suggesting that DPFR's adaptive gating is especially effective at resolving ambiguous TC boundaries.
 
Transformer-based baselines (UNETR, Swin-UNETR) perform comparably to each other in average Dice (0.68) but exhibit substantially worse HD95 performance, with average values of 35.10\,mm and 44.84\,mm respectively. These results suggest that attention-heavy models struggle with boundary precision in the noisy, low-data SSA setting under the evaluated training protocol. SegMamba and SegResNet3D offer improved HD95 over Transformers (16.82\,mm and 30.53\,mm) but remain below MMRINet's boundary precision.

\begin{table}[!t]
\centering
\caption{Segmentation performance on the BraTS-Lighthouse SSA 2025 validation set. DSC and HD95 are reported per subregion and averaged. Best results in \textbf{bold}.}
\label{tab:results}
\begingroup
\small
\renewcommand{\arraystretch}{0.84}
\begin{adjustbox}{width=\columnwidth,center}
\begin{tabular}{lcccccccc}
\toprule
\textbf{Model} & \multicolumn{4}{c}{\textbf{DSC}} & \multicolumn{4}{c}{\textbf{HD95 (mm)}} \\
\cmidrule(lr){2-5} \cmidrule(lr){6-9}
 & ET & TC & WT & AVG & ET & TC & WT & AVG \\
\midrule
UNETR~\cite{unetr}              & 0.64 & 0.62 & 0.77 & 0.68 & 56.83 & 18.14 & 30.33 & 35.10 \\
Swin-UNETR~\cite{swinunetr}     & 0.62 & 0.62 & 0.81 & 0.68 & 33.43 & 40.72 & 60.38 & 44.84 \\
SegMamba~\cite{segmamba}        & 0.62 & 0.64 & 0.82 & 0.69 & 22.79 & 12.55 & 15.11 & 16.82 \\
SegResNet3D~\cite{myronenko20183d} & 0.63 & 0.66 & 0.82 & 0.70 & 62.45 & 12.92 & 16.21 & 30.53 \\
MONAI 3D U-Net~\cite{3dunet}    & 0.68 & 0.71 & 0.80 & 0.73 & \textbf{9.57} & 22.13 & 21.95 & 17.88 \\
MMRINet (Ours)                  & \textbf{0.69} & \textbf{0.73} & \textbf{0.84} & \textbf{0.75} & 16.21 & \textbf{8.12} & \textbf{12.36} & \textbf{12.23} \\
\bottomrule
\end{tabular}
\end{adjustbox}
\endgroup
\end{table}

\subsection{Computational Efficiency}
One of our claims in this work is that strong segmentation performance is achievable without heavy compute. Table~\ref{tab:efficiency} provides a comparison of computational profiles across all evaluated models.
 
\begin{table}[!t]
\centering
\caption{Computational efficiency comparison. Params: trainable parameters (M). Inference time is measured per volume with batch size 1 and FP16 inference on a single NVIDIA RTX A6000 GPU. Peak GPU memory is measured during inference at $160\times160\times128$ resolution.}
\label{tab:efficiency}
\begingroup
\small
\renewcommand{\arraystretch}{0.84}
\begin{tabular}{lccc}
\toprule
\textbf{Model} & \textbf{Params (M)} & \textbf{Inference Time (s)} & \textbf{Peak GPU Mem (GB)} \\
\midrule
SegMamba~\cite{segmamba}          & 18.8 & $\sim$4.2 & $\sim$14.1 \\
UNETR~\cite{unetr}                & 13.2 & $\sim$3.8 & $\sim$11.6 \\
SegResNet3D~\cite{myronenko20183d} & 4.7  & $\sim$2.1 & $\sim$6.8  \\
Swin-UNETR~\cite{swinunetr}       & 4.1  & $\sim$3.5 & $\sim$9.2  \\
MONAI 3D U-Net~\cite{3dunet}      & \textbf{2.0}  & $\sim$\textbf{1.4} & $\sim$\textbf{4.3}  \\
MMRINet (Ours)                    & 2.5  & $\sim$1.8 & $\sim$4.9  \\
\bottomrule
\end{tabular}
\endgroup
\end{table}
 
MMRINet contains only 2.5M parameters, 87\% fewer than SegMamba and 81\% fewer than UNETR, while achieving better segmentation performance on both average Dice and HD95. The MONAI 3D U-Net remains the smallest and fastest baseline, but it trails MMRINet by 0.02 average Dice and 5.65\,mm average HD95. MMRINet's inference time of $\sim$1.8\,s per volume is therefore competitive with this lightweight convolutional baseline, and its peak GPU memory footprint of $\sim$4.9\,GB leaves substantially more deployment headroom than larger transformer and Mamba baselines. For reference, SegMamba requires $\sim$14.1\,GB of VRAM under the same inference protocol; this can fit on some 16\,GB consumer GPUs with careful memory management, but remains less accessible than MMRINet's sub-5\,GB footprint. This efficiency is a direct result of replacing quadratic self-attention with linear Mamba SSM blocks in the bottleneck and of the DPFR design that enriches decoder representations through architectural diversity rather than parameter scaling.

\subsection{Ablation Studies}
 
\subsubsection{Component-Wise Ablation.}
 
Table~\ref{tab:ablation_gains} isolates the contribution of each novel component. Starting from a baseline encoder-decoder with GroupMamba bottleneck (no additional enhancements), each component is added incrementally:
 
\begin{table}[htbp]
\centering
\caption{Incremental component ablation on the BraTS-Lighthouse SSA 2025 validation set.}
\label{tab:ablation_gains}
\begingroup
\small
\renewcommand{\arraystretch}{0.84}
\begin{tabular}{@{}lcc@{}}
\toprule
\textbf{Configuration} & \textbf{DSC (Avg)} & \textbf{HD95 (mm, Avg)} \\
\midrule
Baseline (Conv Bottleneck + U-Net)    & 0.67 & 22.30 \\
Baseline (GroupMamba + U-Net)         & 0.69 & 18.80 \\
+ Deep Supervision                    & 0.71 & 17.69 \\
+ DPFR                                & 0.74 & 17.29 \\
+ DPFR + Deep Supervision             & 0.75 & 14.87 \\
\rowcolor{green!10}
\textbf{Full MMRINet (+ PFA)}         & \textbf{0.75} & \textbf{12.23} \\
\bottomrule
\end{tabular}
\endgroup
\end{table}
 
The DPFR module provides the largest Dice gain in the incremental ablation: adding DPFR to the GroupMamba baseline improves average Dice from 0.69 to 0.74 and reduces HD95 from 18.80\,mm to 17.29\,mm. When combined with deep supervision, DPFR improves the corresponding model from 0.71 to 0.75 Dice and reduces HD95 from 17.69\,mm to 14.87\,mm. Deep supervision contributes a consistent +0.02 Dice improvement by improving gradient flow through the compact network. PFA's contribution is most clearly visible in the HD95 metric: adding PFA to the DPFR + Deep Supervision configuration reduces average HD95 from 14.87\,mm to 12.23\,mm ($-2.64$\,mm), reflecting sharper boundary delineation through multi-scale aggregation without a corresponding change in rounded region-level Dice. The full model improves over the GroupMamba U-Net baseline by +0.06 Dice and 35\% relative HD95 reduction, demonstrating strong interaction between all three components. Replacing the GroupMamba bottleneck with an equivalent-capacity standard 3D convolutional block (Conv Bottleneck row) results in a 0.02 Dice drop and 3.50\,mm HD95 increase, supporting the value of linear-complexity state-space modeling for global context capture.
 
\subsubsection{DPFR Module Analysis.}
 
The DPFR module integrates four sub-components: the detail path (depthwise-separable $3\times3\times3$ conv), the context path (dilated $3\times3\times3$ conv, dilation=2), cross-path interaction ($1\times1\times1$ conv), and adaptive gating. Qualitatively, the detail path is the primary contributor to ET segmentation performance, where fine-grained boundary detection is critical. The context path improves WT delineation by suppressing false positives in high-noise regions surrounding the tumor. Adaptive gating allows the module to dynamically weight these contributions from global feature statistics, functioning as a lightweight path-selection mechanism over the two representational streams.
 
\subsection{Qualitative Results}
 
\begin{figure}[H]
    \centering
    \includegraphics[width=0.44\linewidth,trim={5pt 5pt 5pt 5pt},clip]{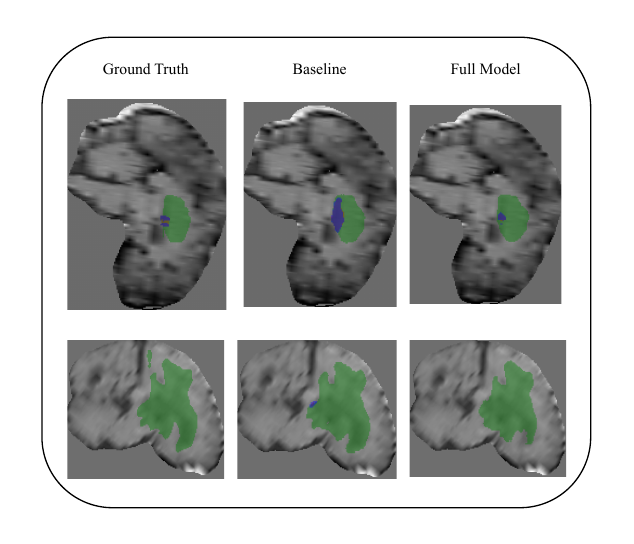}
    \caption{Qualitative segmentation results on BraTS-Lighthouse SSA 2025 validation cases. Green: WT; blue: ET.}
    \label{fig:qualitative}
\end{figure}

Figure~\ref{fig:qualitative} shows representative segmentation outputs from the validation set. MMRINet accurately delineates complex tumor morphologies including irregular boundaries, heterogeneous enhancing regions, and diffuse whole-tumor extents. Compared to the GroupMamba baseline (without DPFR and PFA), the full model shows visibly reduced over-segmentation of the ET subregion and tighter boundary adherence in the WT subregion, consistent with the quantitative HD95 improvements reported in Table~\ref{tab:ablation_gains}. The qualitative panel is intended as a representative visual check; broader best/median/worst-case visualization is left to supplementary material in future releases.
 
\section{Limitations and Future Work}
\label{sec:limitations}
This study is intentionally scoped to the BraTS-Lighthouse SSA 2025 validation split, which matches the target deployment scenario while keeping the evaluation aligned with the challenge protocol. The dataset contains only 60 training cases and 35 validation cases; therefore, the reported improvements should be interpreted as fixed-split validation evidence on a challenging low-resource benchmark. Future work will extend the evaluation to additional cohorts and repeated-seed statistical analysis, broaden consumer-GPU profiling, explore ET-specific loss or post-processing, and expand qualitative analysis to include best, median, worst, and failure cases with all tumor subregions.

\section{Conclusion}
\label{sec:conclusion}
We introduced MMRINet, a lightweight Mamba-based architecture for brain tumor segmentation tailored for resource-limited clinical environments. By integrating linear-complexity state-space modeling with Dual-Path Feature Refinement (DPFR) and Progressive Feature Aggregation (PFA), MMRINet balances contextual awareness and fine structural detail, achieving competitive validation performance (0.75 Dice, 12.23\,mm HD95) with only 2.5M parameters. The model surpasses the evaluated transformer-based and Mamba-based baselines while maintaining a small memory footprint, supporting its effective use as a strong practical segmentation backbone for low-resource MRI analysis.

%
% ---- Bibliography ----
%
% BibTeX users should specify bibliography style 'splncs04'.
% References will then be sorted and formatted in the correct style.
%
% \bibliographystyle{splncs04}
% \bibliography{mybibliography}
%

\end{document}